\documentclass[conference]{IEEEtran}
\IEEEoverridecommandlockouts
\usepackage{cite}
\usepackage{amsmath,amssymb,amsfonts}
\usepackage{graphicx}
\usepackage{textcomp}
\usepackage{xcolor}

\usepackage{graphicx,caption,subcaption}
\usepackage{amsmath,amssymb,stmaryrd,mathtools}
\usepackage[ruled,vlined,titlenotnumbered,linesnumbered]{algorithm2e} 
\usepackage[noend]{algpseudocode}
\usepackage{acronym}
\usepackage{comment}
\usepackage{color}
\usepackage{units}
\usepackage{color, colortbl}
\usepackage{adjustbox}
\usepackage{array,multirow}
\usepackage{xcolor}
\usepackage{soul}
\usepackage{wrapfig}
\usepackage{fontawesome5}

\definecolor{darkGray}{gray}{0.7}
\definecolor{lightGray}{gray}{0.9}

\usepackage[capitalise]{cleveref}

\DeclareCaptionLabelSeparator{periodspace}{.\quad}
\crefname{equation}{}{} 
\crefname{section}{Sec.}{Sec.}

\newcommand{\tvar}{t}
\newcommand{\thor}{T} 

\newcommand{\tinit}{0}

\newcommand{\ctrl}{u}
\newcommand{\dstb}{d}
\newcommand{\cfunc}{u(\cdot)}
\newcommand{\dfunc}{d(\cdot)}
\newcommand{\cset}{\mathcal{U}}

\newcommand{\dset}{\mathcal{D}}

\newcommand{\state}{\mathbf{x}}

\newcommand{\traj}{\xi} 
\newcommand{\dyn}{f} 
\newcommand{\targetfunc}{l}
\newcommand{\targetset}{\mathcal{F}}
\newcommand{\costfunctional}{J}

\newcommand{\vfunc}{V}
\newcommand{\brs}{\mathcal{V}} 

\newcommand{\ham}{H}

\newcommand{\tdummy}{\tau}

\newcommand{\controller}{u}
\newcommand{\ctrlnom}{u_{\text{nom}}}

\newcommand{\stratset}{\Gamma}
\newcommand{\shield}{\text{\scalebox{0.6}{\faShield*}}}

\newcommand{\horizon}{T}

\newcommand{\timestep}{\Delta{t}}

\newcommand{\argmax}{\operatornamewithlimits{argmax}}

\usepackage{lipsum}
\usepackage{url}
\usepackage{amsthm}
\usepackage{multirow}

\begin{document}

\title{Timely Activation of Safety Filters via One-Step Reachability Expansion
}

\author{

\IEEEauthorblockN{Javier Borquez}
\IEEEauthorblockA{
\textit{Universidad de Santiago de Chile}\\
Santiago, Chile \\
javier.borquez@usach.cl}
}

\maketitle

\begin{abstract}
Least-restrictive safety filters based on Hamilton–Jacobi reachability provide strong safety guarantees by overriding a nominal controller only when the system reaches the boundary of the set of unsafe states defined as a Backward Reachable Tube (BRT). These guarantees, however, rely on the continuous-time nature of the underlying formulation. In practice, robotic systems apply control at discrete sampling intervals, which creates a mismatch where the system may “jump” into the unsafe BRT between updates, allowing failures that are theoretically avoidable.
This work introduces a principled solution based on a one-step expanded BRT that predicts all states capable of reaching the true BRT within a single timestep. By using this expanded boundary as the activation condition for the safety filter, safety interventions occur early enough to ensure correctness under discrete-time execution. We formulate this expanded set as a modified reachability problem and compute it using standard continuous-time solvers. 
\end{abstract}
\vspace{1em}
%
\section{\label{intro}Introduction}


As robots and autonomous vehicles are increasingly deployed to accomplish tasks in cluttered environments, it becomes essential find methods to fulfill tasks while avoiding undesirable outcomes. A variety of methods have been proposed to address navigation and path planning under obstacle constraints ~\cite{dolgov2008practical,BADGR2021,chintala2022,swadi2023}. Complementary to these task-oriented strategies, where safety is often treated as a soft constraint, control-theoretic methods provide formal safety guarantees, including Control Barrier Functions (CBFs)~\cite{ames2016control,ames2019control,XU201554}, robust and Tube Model Predictive Control (MPC)~\cite{mayne2011tube,kohler2020computationally}, reachability-based methods~\cite{lygeros2004reachability,mitchell2005time}, and set-based representations such as zonotopes~\cite{Althoff2011zono,Girard2005zono}. These approaches differ in their assumptions and guarantees, yet share the common objective of enabling safe autonomous operation.

Among these methods, Hamilton–Jacobi (HJ) reachability analysis provides a powerful and constructive framework for safety verification and controller synthesis. Unlike most alternatives, HJ reachability directly handles general nonlinear dynamics, bounded controls, and adversarial disturbances within a unified formulation~\cite{bansal2017hamilton,evans1984differential}. Its solutions yields the \emph{Backward Reachable Tube} (BRT), a set capturing all the states from where the system cannot avoid entering a failure state in the future. This formulation not only certifies whether a state is safe but also provides the corresponding optimal safety-preserving control policy.

A common way to leverage the HJ solution in practice is through safe \emph{least-restrictive filtering} (LRF)~\cite{Fisac2023safe_filter,DD_safe_fltr_2023}, where an arbitrary nominal controller governs the system whenever it remains in the safe region, and the optimal safety-preserving control intervenes only when the state reaches the BRT boundary. This approach guarantees safety in the continuous-time setting while preserving nominal behavior elsewhere. The resulting controller is simple, interpretable, and compatible with most existing control or planning algorithms.

A wide range of efficient solvers have been developed for computing continuous-time HJ reachability, including numerical schemes based on level-set methods~\cite{mitchell2004toolbox,helperOC,hj_reach_ASL2023} as well as deep-learning–based approximations~\cite{bansal2021deepreach,feng2025MPC_deepreach}. These approaches make the continuous-time Hamilton–Jacobi–Isaacs variational inequality (HJI-VI) tractable even for moderately high-dimensional systems, providing strong theoretical guarantees and reliable numerical behavior. 

In contrast, direct discrete-time formulations of reachability remain far less developed and are rarely supported by existing toolchains~\cite{MITCHELL2012discreteHJ}. As a result, most practical implementations rely on continuous-time BRTs even though real robotic systems operate with discretized sensing, planning, and control updates. This mismatch introduces a potential safety gap, as the continuous-time guarantees may not strictly hold between discrete control updates, effect which becomes more pronounced as the sampling period increases.


A key consequence of the continuous–discrete mismatch appears when applying LRFs derived from continuous-time BRTs. Within a single control interval $\Delta t$, a system starting outside the BRT may enter it before the filter can react, effectively “jumping” into a doomed state. To mitigate this, practitioners often inflate the BRT by a heuristic margin $\epsilon$, triggering the filter earlier~\cite{borquezFiltering2023,lin2024onefilter}. Although effective, this approach is ad-hoc, requires manual tuning, and can introduce unnecessary conservatism.

In this work, we address the problem of safely implementing LRFs under discrete-time control by introducing a principled method to compute a \emph{One-step Expanded BRT}. This expanded boundary guarantees that the system cannot reach the original BRT within a single sampling interval, ensuring that the safety filter has sufficient time to intervene. The proposed approach is systematically derived from the dynamics and worst-case evolution of the system, preserving the least-restrictive property while minimizing unnecessary conservatism.


\newpage
\section{\label{background}Background}

\subsection{\label{background_hj}Hamilton-Jacobi Reachability}

One way to guarantee safe operation of autonomous  dynamical systems is through Hamilton-Jacobi reachability analysis. This approach involves computing the BRT of a failure set $\targetset$. The BRT captures the states from which the system is not able to avoid entering $\targetset$ within some time horizon $\thor$, despite the best control effort when facing adversarial disturbance. Mathematically the BRT can be defined as:
\begin{equation}\label{eq:brt_set_notation}
\brs(\tvar)=\{\state:\exists\dstb\in\dset,\forall\ctrl\in\cset,\exists\tdummy\in[\tvar,\thor],\traj^{\ctrl,\dstb}_{\state,\tvar}(\tdummy)\in\targetset\}.
\end{equation}
Where $\state$ denotes the state, $\ctrl \in \cset$ the bounded control, and $\dstb \in \dset$ the bounded disturbance, and $\traj$ denoting the trajectory followed by the system under the system dynamics given by $\dot{\state} = \dyn(\state, \ctrl, \dstb)$, in this context the trajectory notation is to be read as the state achieved at time $\tdummy$ by starting at initial state $\state$ and initial time $\tvar$, and applying control and disturbance sequences $\{\ctrl(\cdot),\dstb(\cdot)\}$ over the time interval $[\tvar,\tdummy]$.\\

In HJ reachability, the BRT computation is formulated as a zero-sum game between control and disturbance. Proximity to the failure set $\targetset$ is captured by a signed distance function $\targetfunc(\state)$. Using this function, we express the cost functional as the minimum distance to $\targetset$ along system trajectories over time:
\begin{equation}\label{eq:brt_cost}
\costfunctional(\state, \tvar, \cfunc, \dfunc)=\min _{\tdummy \in[\tvar, \thor]} \targetfunc (\traj^{\ctrl,\dstb}_{\state,\tvar}(\tdummy)).
\end{equation}
The goal is to capture this minimum distance for optimal system trajectories. Thus, we compute the optimal control that maximizes this distance (drives the system away from the failure set) and the worst-case disturbance signal that minimizes the distance. The value function corresponding to this robust optimal control problem is:
\begin{equation}\label{eq:hji}
 \vfunc(\state, \tvar)=\adjustlimits\inf_{\dstb \in \stratset(\tvar)} \sup_{\cfunc} \{\costfunctional(\state, \tvar, \cfunc, \dstb[\ctrl](\cdot))\},
\end{equation}
where $\stratset(t)$ defines the set of non-anticipative strategies for the disturbance \cite{bansal2017hamilton}.
The value function in (\ref{eq:hji}) can be computed using dynamic programming, which results in the following final value Hamilton-Jacobi-Isaacs Variational Inequality (HJI-VI) \cite{bansal2017hamilton,lygeros2004reachability,mitchell2005time}:
\begin{equation} \label{eq:pde}\fontsize{9.5}{10}\selectfont
    \begin{aligned}
    \min \{D_{\tvar} \vfunc(\state, \tvar)& + \ham(\state, \tvar, \nabla \vfunc(\state, \tvar)), \targetfunc(\state) - \vfunc(\state, \tvar) \} = 0, \\
    &\vfunc(\state, \horizon) = \targetfunc(\state), \quad \text{for} \ \tvar \in \left[\tinit, \horizon\right].
    \end{aligned}
\end{equation}
$D_t$ and $\nabla$ represent the time and spatial gradients of the value function. $\ham$ is the Hamiltonian, which optimizes over the inner product between the spatial gradients of the value function and the dynamics:
\begin{equation}\label{eq:ham_safe}
    \ham(\state, \tvar, \nabla \vfunc(\state, \tvar)) = \max_{\ctrl \in \cset} \min_{\dstb \in \dset} \nabla \vfunc(\state, \tvar) \cdot \dyn(\state, \ctrl, \dstb).
\end{equation}
%
Once the value function is obtained, the BRT is given as the sub-zero level set of the value function:
\begin{equation}
\brs(\tvar)=\{\state: \vfunc(\state, \tvar) \leq 0\}.
\end{equation}
The corresponding optimal safe control can be derived as :
\begin{equation}\label{eq:optctrl_safe}
\ctrl^{\shield}(\state, \tvar)=\argmax_{\ctrl \in \cset} \min_{\dstb \in \dset}\nabla \vfunc(\state, \tvar) \cdot \dyn(\state, \ctrl, \dstb).
\end{equation}
%


In safety-ensuring applications, we want to guarantee that the system does not enter $\targetset$ \textit{at any time}. To achieve this, we use a time-converged BRT, as the set of unsafe states typically stops expanding after some time. Since our objective in this work is safety, we will rely exclusively on the converged value function $V(\state)$ throughout this paper. This allows us to synthesize safety controllers using an expression identical to (\ref{eq:optctrl_safe}) but without time dependency.

\subsection{\label{background_lr}Least Restrictive Filtering}

If the BRT and associated converged value function are known, a \textit{Least Restrictive Filter} (LRF) can be deployed to guarantee the safe operation of the system.
The LRF is constructed as follows:
\begin{equation}\label{eq:lst_restrict_safety_ctrl}
\controller(\state, \tvar) = \begin{cases}
  \ctrlnom(\state, \tvar) & \vfunc(\state)> 0, \\
   \ctrl^{\shield}(\state) & \vfunc(\state) = 0.
\end{cases}
\end{equation}
Here, $\ctrlnom(\state, \tvar)$ corresponds to an arbitrary nominal controller that might optimize a performance criterion without enforcing safety constraints. This control is used when the value function $\vfunc(\state)$ is positive, as the system is not at risk of breaching safety. Whenever the system reaches the boundary of the BRT ($\vfunc(\state)=0$), it switches to $\ctrl^{\shield}(\state)$ as this optimal control is guaranteed to maintain or increase $\vfunc(\state)$ keeping the system from entering the unsafe states determined by the BRT, thereby enforcing safety at all times. We refer the reader to \cite{borquezFiltering2023,Fisac2023safe_filter} for details on this filtering technique and proof of the safety guarantees.

\section{\label{problem}Problem Statement}

The least-restrictive filter in \eqref{eq:lst_restrict_safety_ctrl} guarantees safety in continuous time as the control can be switched instantaneously when $\vfunc(\state)=0$. However, in practice, robotic systems operate in discrete time with a finite sampling period $\timestep$. This introduces a mismatch between the continuous-time safety guarantees provided by the HJ formulation and the discrete-time execution of control actions.

Let $\brs = \{ \state : \vfunc(\state) \leq 0 \}$ denote the time-converged Backward Reachable Tube corresponding to the unsafe set $\targetset$. When control updates occur at discrete intervals $\timestep$, the system may transition from a state $\state_t$ with $\vfunc(\state_t) > 0$ (nominal control region) to a state $\state_{t+\timestep}$ satisfying $\vfunc(\state_{t+\timestep}) < 0$ making the safety filter react from an unsafe state. This leads to inevitable safety violations despite the use of the optimal safety filter.

\begin{figure}[t] 
\begin{center} 
\vspace{0.0em}
\includegraphics[width=0.99\columnwidth]{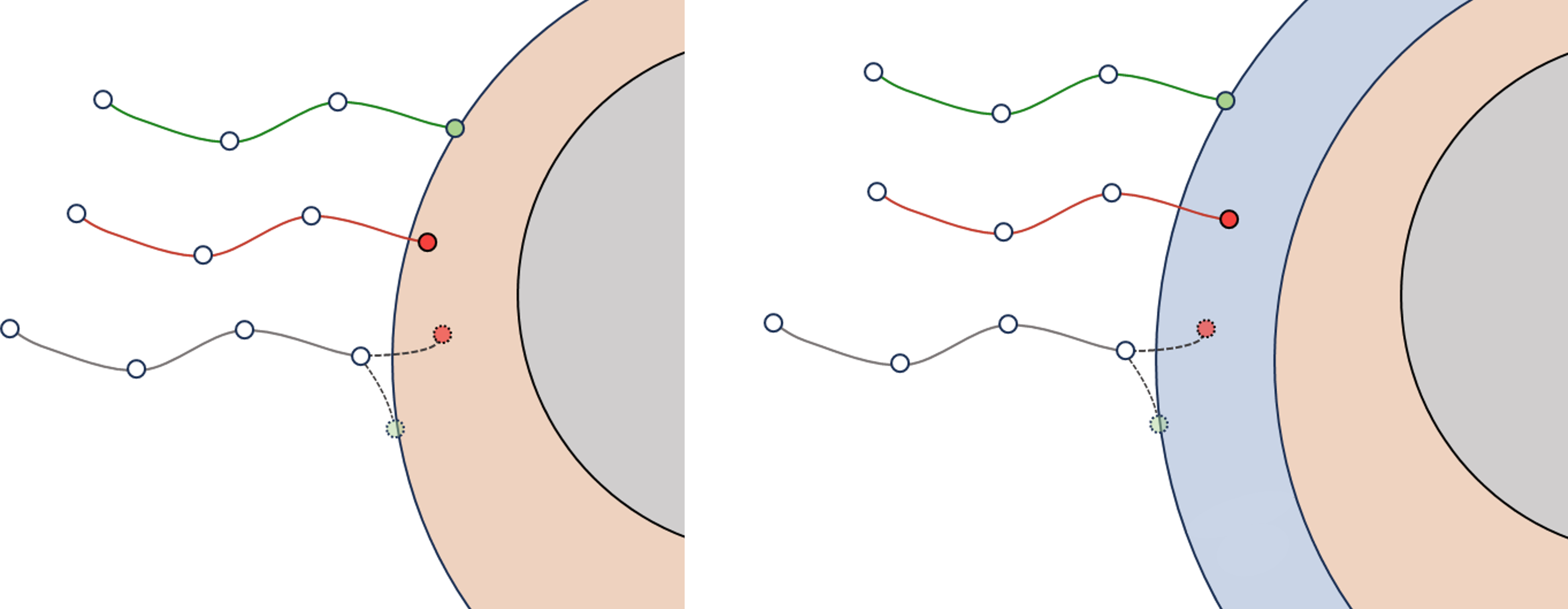}
\vspace{-1.5em}
\caption{Effect of control discretization on safety. Grey denotes the obstacle set and light orange the BRT. (Left) Green: discrete update at the boundary maintains safety. Red: update inside the BRT guarantees a violation. Black: update before the boundary applies the nominal policy, which may still lead to failure. (Right) Blue: inflated BRT boundary requiring at least one $\Delta t$ to traverse. Green: update at the inflated boundary maintains safety. Red: update inside the inflated boundary allows the safety filter to activate before reaching the true BRT boundary, preserving safety. Black: update before the inflated boundary applies the nominal policy, which remains safe as the system cannot move from outside the inflated boundary to inside the BRT within a single update step.}
\vspace{0.0em}
\label{fig:dt_problem_and_sol}
\end{center}
\end{figure}

To avoid this issue, we can rely on an inflated BRT as the activation boundary for the LRF in \eqref{eq:lst_restrict_safety_ctrl} that expands the original boundary by some margin, such as the common heuristic of using a sufficiently big $\epsilon$ super-level of the value function. The key requirement for any successful expansion is that any state on the boundary of the inflated BRT must require more than one sampling interval $\timestep$ to reach the original BRT under worst-case dynamics. Intuitively, this ensures that the system cannot transition from outside the inflated boundary to inside the true BRT within a single discrete control step, effectively eliminating the safety violations introduced by control discretization.
This is visually presented in Figure~\ref{fig:dt_problem_and_sol}.

The core challenge is to determine, for a given sampling period $\timestep$, the minimal expansion of the base BRT that guarantees safety under discrete-time control updates. The goal is to construct an inflated BRT that serves as the activation boundary for the least-restrictive filter, ensuring that the system always triggers the safety intervention before crossing into the original BRT during one control interval. This approach provides a principled way to compensate for the effects of control discretization in the LRF while preserving its least-restrictive nature.
\section{\label{approach}Proposed Approach}

\subsection{\label{subsec:expandedBRT}One-step Expanded BRT}

To formalize the notion of a minimal expansion consistent with a given control update period $\timestep$, we introduce the One-step Expanded BRT. This set captures all states that could reach the original BRT within a single sampling interval under the system’s worst-case evolution.
\begin{equation}\fontsize{9.5}{10}\selectfont
\brs_{\timestep}(\tvar)=\{\state:\exists\dstb\in\dset,\exists\ctrl\in\cset,\exists\tdummy\in[\tvar,\timestep],\traj^{\ctrl,\dstb}_{\state,\tvar}(\tdummy)\in\brs\}.
\end{equation}
Unlike the standard formulation in \eqref{eq:brt_set_notation}, this definition treats the base, time-converged BRT as the new failure set to avoid, with a time horizon of $\timestep$. The control and disturbance inputs are both assumed to act adversarially to safety, capturing the most aggressive evolution of the system toward the unsafe region. This models the worst-case behavior that could occur during the interval where the nominal controller remains active.

Based on this definition, we introduce corresponding cost and value functions, analogous to \eqref{eq:brt_cost} and \eqref{eq:hji}, respectively:
\begin{equation}\label{eq:brt_cost_expanded}
\costfunctional_{\timestep}(\state, \tvar, \cfunc, \dfunc)=\min _{\tdummy \in[\tvar, \timestep]} \vfunc(\traj^{\ctrl,\dstb}_{\state,\tvar}(\tdummy)),
\end{equation}
\begin{equation}\label{eq:hji_expanded}
 \vfunc_{\timestep}(\state, \tvar)=\adjustlimits\inf_{\dstb \in \stratset(\tvar)} \inf_{\cfunc} \{\costfunctional_{\timestep}(\state, \tvar, \cfunc, \dstb[\ctrl](\cdot))\}.
\end{equation}
These functions describe the minimal value of $\vfunc$ that can be reached under worst-case dynamics within a single sampling period $\timestep$. In this way, $\vfunc_{\timestep}$ characterizes the minimal distance to the true base BRT achievable in one discrete control step.

This formulation naturally extends the standard HJI-VI in \eqref{eq:pde} to compute the One-step Expanded BRT:
\begin{equation} \label{eq:pde_expanded}\fontsize{9}{10}\selectfont
    \begin{aligned}
    \min \{D_{\tvar} \vfunc_{\timestep}(\state, \tvar)& + \ham_{\timestep}(\state, \tvar, \nabla \vfunc_{\timestep}(\state, \tvar)), \vfunc(\state) - \vfunc_{\timestep}(\state, \tvar) \}\hspace{-0.3em}=\hspace{-0.2em} 0, \\
    &\vfunc_{\timestep}(\state, \timestep) = \vfunc(\state), \quad \text{for} \ \tvar \in \left[\tinit, \timestep\right].
    \end{aligned}
\end{equation}
Here, the terminal condition ensures that at time $\timestep$, the inflated value function matches the base value function $\vfunc(\state)$. The evolution of $\vfunc_{\timestep}$ backward in time over the interval $\tvar \in [\tinit, \timestep]$ captures the region from which the system can be driven into the original BRT within one sampling period.

Finally, the modified Hamiltonian $\ham_{\timestep}$ encodes the adversarial actions for control and disturbance that most rapidly drives the system toward the base BRT:
\begin{equation}\label{eq:HJIVI_ham_expanded}\fontsize{9}{10}\selectfont
    \ham_{\timestep}(\state, \tvar, \nabla \vfunc_{\timestep}(\state, \tvar)) = \min_{\ctrl \in \cset} \min_{\dstb \in \dset} \nabla \vfunc_{\timestep}(\state, \tvar) \cdot \dyn(\state, \ctrl, \dstb).
\end{equation}
This form ensures that the inflated BRT represents the set of states that within one full $\timestep$ can reach the original BRT under worst-case conditions.
A full derivation of \eqref{eq:pde_expanded} follows by starting from the one-step value function definition in \eqref{eq:hji_expanded}, applying the dynamic programming principle, and deriving the corresponding HJI variational inequality and Hamiltonian. For brevity, the detailed derivation is omitted, as it directly parallels the standard HJI-VI development with the role of the control in the optimization reversed.
Once the value function is obtained, the One-step Expanded BRT is given as the sub-zero level set of the value function:
\begin{equation}
\brs_{\timestep}(\tvar)=\{\state: \vfunc_{\timestep}(\state, \tvar) \leq 0\}.
\end{equation}
This formulation allows the use of standard HJ reachability solvers to compute the one-step inflated BRT directly. By defining both the control and disturbance inputs to act adversarially to safety (minimizing the value function) within the solver, and by setting the converged value function $\vfunc(\state)$ as the terminal condition, we can propagate the HJI-VI in \eqref{eq:pde_expanded} backward for one sampling interval $\timestep$ to obtain $\vfunc_{\timestep}$.

\subsection{\label{proof}Safety of the One-step Expansion}

In this subsection we show that activating the LRF using the inflated BRT ensures that safety is maintained under discrete-time control updates.

\textbf{Proposition 1.}
If the least-restrictive safety filter activates using as switching condition the boundary of the one-step inflated BRT $\brs_{\timestep}$, that is:

\begin{equation}\label{eq:lr_expandedBRT}
\controller(\state, \tvar) = \begin{cases}
  \ctrlnom(\state, \tvar) & \vfunc_{\timestep}(\state)> 0, \\
   \controller^{\shield}(\state) & \vfunc_{\timestep}(\state) \leq 0.
\end{cases}
\end{equation}
then safety is guaranteed for discrete-time control execution with period $\timestep$ for any state starting outside of $\brs_{\timestep}$.

\begin{proof}
Let $\state_t$ be the state at a sampling instant $t$. Assume $\state_t \notin \brs_{\timestep}$, i.e. $\vfunc_{\timestep}(\state_t) > 0$. The control law \eqref{eq:lr_expandedBRT} implies the nominal controller $\ctrl_{\text{nom}}$ is applied on the interval $[t, t+\timestep)$. Denote by $\state_{t+\timestep}=\traj^{\ctrl_{\text{nom}},\dstb}_{\state_t,t}(\timestep)$ the state reached after one sampling period under some (possibly adversarial) disturbance $\dstb(\cdot)$ and the nominal control $\ctrl_{\text{nom}}(\cdot)$.

Consider the mutually exclusive possibilities for the sampled successor $\state_{t+\timestep}$:

\begin{enumerate}

\item $\vfunc_{\timestep}(\state_{t+\timestep})>0$. 
Safety is preserved as this state is outside both the One-step Expanded BRT and base BRT.

\item $\vfunc_{\timestep}(\state_{t+\timestep})\le 0$ and $\vfunc(\state_{t+\timestep})>0$. In this case the LRF will switch to the safe controller at time $t+\timestep$. Since $\state_{t+\timestep}$ is outside the base BRT, the safe controller can keep the trajectory out of $\brs$ thereafter \cite{borquezFiltering2023}.

\item $\vfunc(\state_{t+\timestep})\le 0$. This case is not possible under the assumption $\vfunc_{\timestep}(\state_t)>0$: by construction if $\vfunc_{\timestep}(\state_t)>0$ there is no one-step trajectory that can reach a state with $\vfunc(\state_{t+\timestep})>0$ for all admissible controls and disturbances.

\end{enumerate}

Therefore, starting from any $\state_t$ with $\vfunc_{\timestep}(\state_t)>0$, the system cannot enter the true BRT compromising the system safety. This because after the system evolves during one sampling step the LRF \eqref{eq:lr_expandedBRT} either: (i) uses the nominal policy while safety is preserved by staying outside the One-step Expanded BRT, or (ii) the One-step Expanded BRT is breached and the optimally safe policy is used before the system can reach the true BRT. Hence safety is maintained for discrete-time control execution with period $\timestep$.
\end{proof}

\section{\label{simulations}Simulation Studies}

For a simulation study we consider an autonomous robot modeled as a Dubins' car trying to reach a goal in a cluttered environment. The system can be represented by the following three dimensional dynamics:
\begin{equation}\label{eq:dyn_car}
\dot{\state}
= \begin{bmatrix} \dot{p_x} \\ \dot{p_y} \\ \dot{\theta} \end{bmatrix}
= \begin{bmatrix} V \cos(\theta) + \dstb_x \\ V \sin(\theta) + \dstb_y \\ \ctrl \end{bmatrix},
\end{equation}
where $p_x,p_y$ denote the car’s position, $\theta$ its orientation, $V = 0.3\text{m/s}$ its fixed speed, $\ctrl \in [-0.75, 0.75]\text{rad/s}$ its angular velocity input, and $\dstb = [\dstb_x, \dstb_y]$ a random bounded velocity disturbance satisfying $\|\dstb\|_2 \le 0.03\text{m/s}$.

\begin{figure}[t] 
\begin{center} 
\vspace{0.0em}
\includegraphics[width=0.775\columnwidth]{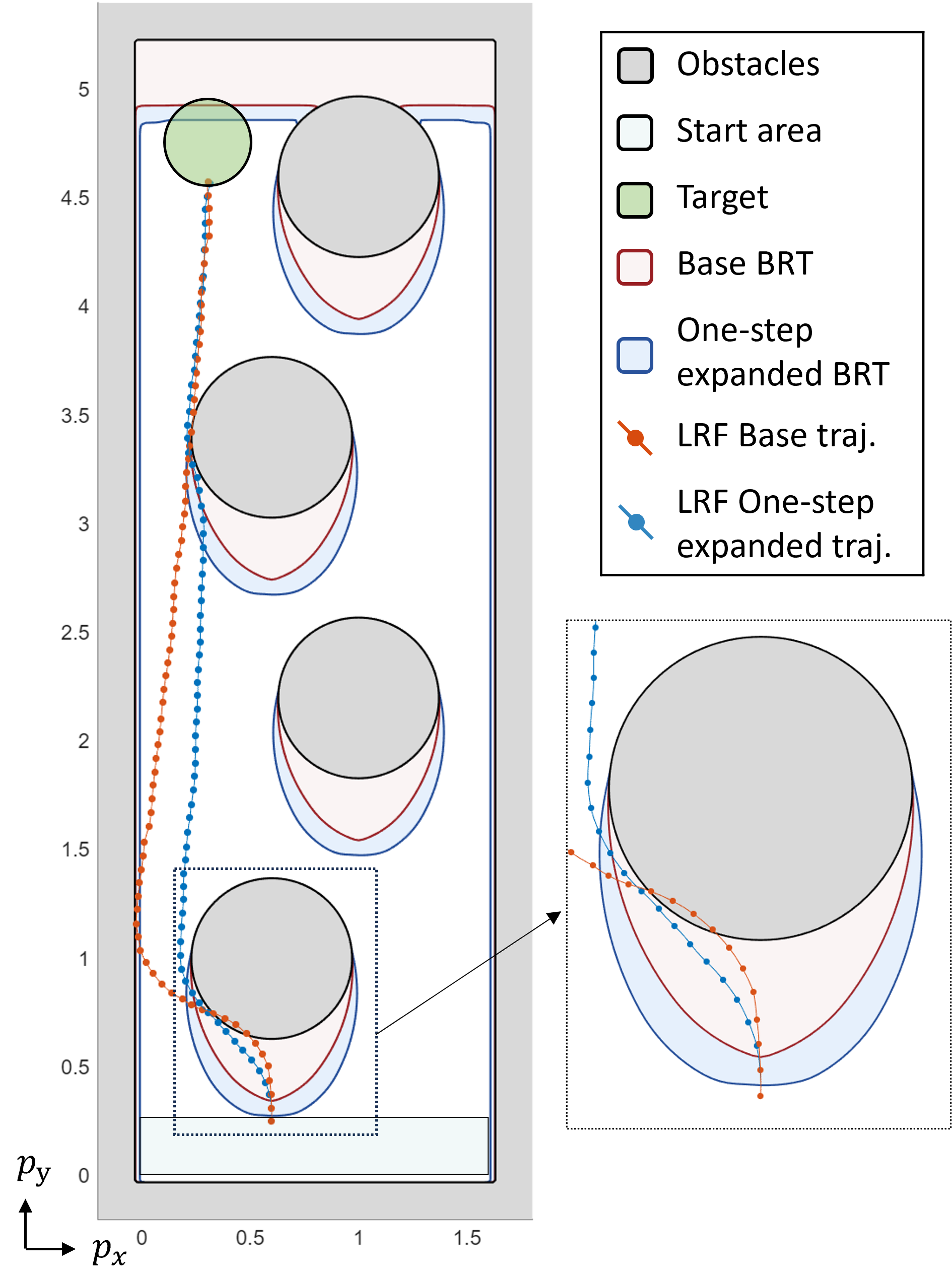}
\vspace{-0.5em}
\caption{One-step Expanded BRT and base BRT for the obstacle set in the simulation environment for Dubins' car study using $\timestep=0.2\text{s}$. Trajectories for each LRF scheme are shown as lines with dots, with dots representing the discrete steps where the control is updated. Here we observe how the early activation of the LRF by the One-step Expanded BRT is crucial in preserving safety in contrast to the base BRT.}
\vspace{-0.5em}
\label{fig:dubins_traj}
\end{center}
\end{figure}

The failure set is defined as a $2\text{m} \times 5.6\text{m}$ rectangular enclosure containing four circular obstacles of diameter $0.2\text{m}$. This set is inflated by $0.17\text{m}$ to account for the robot’s body diameter. The robot’s objective is to travel from a designated start region to a target set located on opposite sides of the enclosure, as shown in Fig.~\ref{fig:dubins_traj} (left).

The base BRT for this setup is computed on a $[p_x,p_y,\theta]$ grid of $[101, 281, 181]$ points , corresponding to a spatial resolution of $2\text{cm}$ and an angular resolution of $2^\circ$. The computation proceeds until convergence, which in this case occurs after a time horizon of $1\text{s}$.

To evaluate the effect of discrete control updates, we simulate safe navigation using four discrete control timesteps, $\timestep \in [0.1, 0.2, 0.3, 0.4]\text{s}$. For each timestep, the One-step Expanded BRT is computed following the procedure described in subsection~\ref{subsec:expandedBRT}, using the same state-space grid as the base BRT. Figure~\ref{fig:dubins_traj} shows both the base BRT and the One-step Expanded BRT for $\timestep = 0.2\text{s}$.
It is worth mentioning that these correspond to 2-dimensional projections of the 3-dimensional BRTs, here we show the projections for $\theta=\pi/2$. 
 
To attempt to complete the task, we use a shooting-based MPC nominal controller that does not consider obstacle avoidance in its objective.
We choose this unsafe nominal controller on purpose to highlight the effect of safety filtering.
The nominal controller will be filtered using the least restrictive filters \eqref{eq:lst_restrict_safety_ctrl} and \eqref{eq:lr_expandedBRT} which use the base BRT and One-step Expanded BRT correspondingly.

Figure~\ref{fig:dubins_traj} shows two trajectories for the filtered nominal policy, each one corresponding to an activation based on the use of the base BRT or One-step Expanded BRT. The left zoomed-in image highlights how the early activation of the safety filter produced by the One-step Expanded BRT allows the system to maintain safety around the obstacle while the late activation in the base BRT case leads to collision in the form of penetration of the obstacle.

For each timestep setting and each filtering method, we perform 100 Monte Carlo simulations with randomly sampled initial states within the starting area. We assume perfect state estimation and an exact dynamics model to isolate as much as possible the impact of delayed safety filter activation from other sources of error. For each run, we report the number of safe trajectories, along with the median of the maximum penetration into the obstacle set, and time to first safety violation.\footnote{Because penetration or failure can occur over a continuous spectrum when failure happens, and remains undefined for successful runs, the resulting distributions of failure-related metrics are heavily skewed. Consequently, the median offers a more representative measure of central tendency than the mean and enables more reliable statistical significance testing.}

\begin{table}[]\centering
\caption{Simulation results for the Dubins' car study under different discrete control timesteps. Metrics report the median across 100 runs.}
\label{tab:dt_results}
\begin{tabular}{|p{.5cm}|p{.8cm}|c|c|c|}
\hline
\multirow{2}{*}{$\timestep$} & \multirow{2}{*}{BRT} & Safe & Maximum & Time to\\
  &  & trajectories    & penetration [mm] & first collision [s]  \\ \hline
\hline
\multirow{2}{*}{$0.1\text{s}$}& Base&11 & $15.7$ & $5.7$ \\
                              & Exp.&67 & $7.5$  & $6.4$ \\ \hline
\multirow{2}{*}{$0.2\text{s}$}& Base&5  & $27.8$ & $5.6$ \\
                              & Exp.&82 & $5.7$  & $6.5$ \\ \hline
\multirow{2}{*}{$0.3\text{s}$}& Base&6  & $35.7$ & $5.4$ \\
                              & Exp.&85 & $16.3$ & $6.4$ \\ \hline
\multirow{2}{*}{$0.4\text{s}$}& Base&6  & $46.1$ & $5.4$ \\
                              & Exp.&86 & $12.1$ & $11.3$ \\ \hline
\end{tabular}
\end{table}

Table~\ref{tab:dt_results} summarizes the results of the simulation study across different control timesteps. The proposed One-step Expanded BRT safety filter consistently outperforms the baseline approach in all evaluated metrics. While the Base BRT filter results in frequent and severe collisions, achieving only 5–11 safe trajectories out of 100 runs, the One-step Expanded BRT maintains substantially higher safety rates, ranging from 67 to 89 across timesteps. The median penetration depths are consistently lower, and failures, when they occur, take longer to manifest. Together, these results indicate that the One-step Expanded BRT not only prevents most collisions but also ensures that when failures do occur, they are milder and occur later in time.

To assess the statistical significance of these differences, one-sided Mann–Whitney U tests were conducted comparing the distributions of both maximum penetration and time to first collision across methods at each $\timestep$. In all cases, the One-step Expanded BRT filter achieved significantly better outcomes with $p < 0.05$, confirming that the observed improvements are not due to random variation. These findings validate that using the One-step Expanded BRT as the activation boundary for least-restrictive filtering provides a principled and effective mechanism to improve safety under discrete-time control execution.

While the proposed method substantially improves safety performance, it does not provide guarantees, as indicated by the nonzero number of failures across all tested timesteps. This residual risk arises from the underlying mismatch between the continuous-time BRT used in the safety filter and the discrete-time execution of control actions. Although the one-step expansion effectively compensates for delayed filter activation caused by control discretization, it still relies on a value function computed under the assumption of continuous time control authority. As a result, the derived safety boundary may not perfectly capture the true discrete-time reachable set, leaving a gap through which safety violations can still occur.

It is worth mentioning that results show improvement in safety performance for larger control timesteps, with a pronounced jump in number of successful trajectories at $\Delta t = 0.2\text{s}$ compared to $\Delta t = 0.1 \text{s}$. One likely explanation is that the One-step Expanded BRT provides stronger safety margins as the timestep grows. Since the expansion is computed under worst-case assumptions for both control and disturbances, while the nominal shooting MPC policy behaves non-adversarially, thus the resulting inflated boundary offers more protection than strictly necessary to address the delayed activation of the filter. This additional margin compensates for residual effects of the mismatch between the continuous-time safety certificate and its discrete implementation, leading to improved robustness at larger sampling intervals.
\section{\label{conclusion}Conclusion}


This work addressed a key gap that arises when continuous-time Hamilton–Jacobi safety certificates are deployed on discrete-time robotic systems. While least-restrictive safety filters guarantee safety when paired with continuous-time Backward Reachable Tubes, these guarantees weaken when control is updated at finite sampling intervals. To mitigate the resulting “jump-over” failures, we introduced the One-step Expanded BRT, derived by solving an adversarial one-step reachability problem. This construction provides the smallest expanded boundary that ensures a system cannot transition from a safe state to inside the unsafe BRT within a single control period.

Through simulated navigation experiments on a Dubins vehicle, we demonstrated that the proposed one-step expanded BRT substantially improves the reliability of least-restrictive filtering across a range of sampling times. The expanded filter produced significantly more safe trajectories, exhibited much smaller penetration values when failures occurred, and consistently delayed the time of first violation compared to the standard LR filter. These improvements were statistically validated using Mann–Whitney U tests on maximum penetration and time-to-first-collision metrics, confirming that the expanded BRT provides meaningful and principled safety advantages.

Although the approach improves safety when using continuous-time BRTs for safety filtering in discrete-time settings, it does not completely eliminate failures. Because the underlying safety certificate is derived in continuous time, the system remains subject to the broader continuous–discrete mismatch, which the one-step expansion only partially resolves. Further mitigating this mismatch could involve constructing safety sets directly in discrete time, or combining continuous-time HJ analysis with conservative discrete-time over-approximations of system flows. Additional future directions include extending the proposed formulation to higher-dimensional systems and validating its effectiveness on real robotic platforms. Developing such approaches remains an important challenge for ensuring robust safety guarantees under realistic control-loop timings.

More broadly, while our method was formulated within the HJ reachability framework, nothing restricts it to this setting. Any barrier certificate that defines an unsafe set and propagates trajectories under bounded inputs could be expanded using the same adversarial one-step procedure. This includes control barrier functions, tube-model predictive control safety sets, and other viability-based certificates. Developing such generalized expansions, along with discrete-time reachability tools and real-world validation, represents a promising direction for future work.

\bibliographystyle{IEEEtran}
\bibliography{  ./Bib/bib_reach,
                ./Bib/bib_SIAL,
                ./Bib/bib_MPC,
                ./Bib/bib_cbf,
                ./Bib/bib_local,
                ./Bib/bib_fltr
}

\begin{thebibliography}{10}
\providecommand{\url}[1]{#1}
\csname url@samestyle\endcsname
\providecommand{\newblock}{\relax}
\providecommand{\bibinfo}[2]{#2}
\providecommand{\BIBentrySTDinterwordspacing}{\spaceskip=0pt\relax}
\providecommand{\BIBentryALTinterwordstretchfactor}{4}
\providecommand{\BIBentryALTinterwordspacing}{\spaceskip=\fontdimen2\font plus
\BIBentryALTinterwordstretchfactor\fontdimen3\font minus \fontdimen4\font\relax}
\providecommand{\BIBforeignlanguage}[2]{{%
\expandafter\ifx\csname l@#1\endcsname\relax
\typeout{** WARNING: IEEEtran.bst: No hyphenation pattern has been}%
\typeout{** loaded for the language `#1'. Using the pattern for}%
\typeout{** the default language instead.}%
\else
\language=\csname l@#1\endcsname
\fi
#2}}
\providecommand{\BIBdecl}{\relax}
\BIBdecl

\bibitem{dolgov2008practical}
D.~Dolgov, S.~Thrun, M.~Montemerlo, and J.~Diebel, ``Practical search techniques in path planning for autonomous driving,'' \emph{ann arbor}, vol. 1001, no. 48105, pp. 18--80, 2008.

\bibitem{BADGR2021}
G.~Kahn, P.~Abbeel, and S.~Levine, ``Badgr: An autonomous self-supervised learning-based navigation system,'' \emph{IEEE Robotics and Automation Letters}, vol.~6, no.~2, pp. 1312--1319, 2021.

\bibitem{chintala2022}
P.~Chintala, R.~Dornberger, and T.~Hanne, ``Robotic path planning by q learning and a performance comparison with classical path finding algorithms,'' \emph{International Journal of Mechanical Engineering and Robotics Research}, vol.~11, no.~6, pp. 373--378, 2022.

\bibitem{swadi2023}
S.~M. Swadi, A.~K. Kadhim, and G.~M. Ali, ``Design of path planning controller of autonomous wheeled mobile robot based on triple pendulum behaviour,'' \emph{Int. J. Mech. Eng. Robot. Res}, vol.~12, no.~1, pp. 23--31, 2023.

\bibitem{ames2016control}
A.~D. Ames, X.~Xu, J.~W. Grizzle, and P.~Tabuada, ``Control barrier function based quadratic programs for safety critical systems,'' \emph{IEEE Transactions on Automatic Control}, 2016.

\bibitem{ames2019control}
A.~D. Ames, S.~Coogan, M.~Egerstedt, G.~Notomista, K.~Sreenath, and P.~Tabuada, ``Control barrier functions: Theory and applications,'' in \emph{IEEE European control conference (ECC)}, 2019.

\bibitem{XU201554}
\BIBentryALTinterwordspacing
X.~Xu, P.~Tabuada, J.~W. Grizzle, and A.~D. Ames, ``Robustness of control barrier functions for safety critical control.'' \emph{IFAC-PapersOnLine}, vol.~48, no.~27, pp. 54--61, 2015, analysis and Design of Hybrid Systems ADHS. [Online]. Available: \url{https://www.sciencedirect.com/science/article/pii/S2405896315024106}
\BIBentrySTDinterwordspacing

\bibitem{mayne2011tube}
D.~Q. Mayne, E.~C. Kerrigan, E.~Van~Wyk, and P.~Falugi, ``Tube-based robust nonlinear model predictive control,'' \emph{International journal of robust and nonlinear control}, vol.~21, no.~11, pp. 1341--1353, 2011.

\bibitem{kohler2020computationally}
J.~K{\"o}hler, R.~Soloperto, M.~A. M{\"u}ller, and F.~Allg{\"o}wer, ``A computationally efficient robust model predictive control framework for uncertain nonlinear systems,'' \emph{IEEE Transactions on Automatic Control}, vol.~66, no.~2, pp. 794--801, 2020.

\bibitem{lygeros2004reachability}
J.~Lygeros, ``On reachability and minimum cost optimal control,'' \emph{Automatica}, vol.~40, no.~6, pp. 917--927, 2004.

\bibitem{mitchell2005time}
I.~Mitchell, A.~Bayen, and C.~J. Tomlin, ``A time-dependent {Hamilton-Jacobi} formulation of reachable sets for continuous dynamic games,'' \emph{IEEE Transactions on Automatic Control (TAC)}, 2005.

\bibitem{Althoff2011zono}
M.~Althoff and B.~H. Krogh, ``Zonotope bundles for the efficient computation of reachable sets,'' in \emph{2011 50th IEEE Conference on Decision and Control and European Control Conference}, 2011, pp. 6814--6821.

\bibitem{Girard2005zono}
A.~Girard, ``Reachability of uncertain linear systems using zonotopes,'' in \emph{Hybrid Systems: Computation and Control}, M.~Morari and L.~Thiele, Eds.\hskip 1em plus 0.5em minus 0.4em\relax Berlin, Heidelberg: Springer Berlin Heidelberg, 2005, pp. 291--305.

\bibitem{bansal2017hamilton}
S.~Bansal, M.~Chen, S.~Herbert, and C.~J. Tomlin, ``{Hamilton-Jacobi Reachability}: A brief overview and recent advances,'' in \emph{IEEE Conference on Decision and Control (CDC)}, 2017.

\bibitem{evans1984differential}
L.~C. Evans and P.~E. Souganidis, ``Differential games and representation formulas for solutions of hamilton-jacobi-isaacs equations,'' \emph{Indiana University mathematics journal}, vol.~33, no.~5, pp. 773--797, 1984.

\bibitem{Fisac2023safe_filter}
K.-C. Hsu, H.~Hu, and J.~F. Fisac, ``The safety filter: A unified view of safety-critical control in autonomous systems,'' \emph{Annual Review of Control, Robotics, and Autonomous Systems}, vol.~7.

\bibitem{DD_safe_fltr_2023}
K.~P. Wabersich, A.~J. Taylor, J.~J. Choi, K.~Sreenath, C.~J. Tomlin, A.~D. Ames, and M.~N. Zeilinger, ``Data-driven safety filters: Hamilton-jacobi reachability, control barrier functions, and predictive methods for uncertain systems,'' \emph{IEEE Control Systems Magazine}, vol.~43, no.~5, pp. 137--177, 2023.

\bibitem{mitchell2004toolbox}
I.~Mitchell, ``A toolbox of level set methods,'' \emph{http://www. cs. ubc. ca/mitchell/ToolboxLS/toolboxLS. pdf, Tech. Rep. TR-2004-09}, 2004.

\bibitem{helperOC}
M.~Chen and S.~Herbert, ``{helperOC Library, 2019},'' \url{https://github.com/HJReachability/helperOC}.

\bibitem{hj_reach_ASL2023}
\BIBentryALTinterwordspacing
E.~Schmerling and M.~Pavone, ``hj reachability: Hamilton-jacobi reachability analysis in jax,'' 2023. [Online]. Available: \url{https://github.com/StanfordASL/hj_reachability}
\BIBentrySTDinterwordspacing

\bibitem{bansal2021deepreach}
S.~Bansal and C.~J. Tomlin, ``{DeepReach}: A deep learning approach to high-dimensional reachability,'' in \emph{IEEE International Conference on Robotics and Automation (ICRA)}, 2021.

\bibitem{feng2025MPC_deepreach}
Z.~Feng, L.~Qiu, and S.~Bansal, ``{Bridging Model Predictive Control and Deep Learning for Scalable Reachability Analysis},'' in \emph{Proceedings of Robotics: Science and Systems}, LosAngeles, CA, USA, June 2025.

\bibitem{MITCHELL2012discreteHJ}
\BIBentryALTinterwordspacing
I.~M. Mitchell, M.~Chen, and M.~Oishi, ``Ensuring safety of nonlinear sampled data systems through reachability1,'' \emph{IFAC Proceedings Volumes}, vol.~45, no.~9, pp. 108--114, 2012, 4th IFAC Conference on Analysis and Design of Hybrid Systems. [Online]. Available: \url{https://www.sciencedirect.com/science/article/pii/S1474667015371822}
\BIBentrySTDinterwordspacing

\bibitem{borquezFiltering2023}
J.~Borquez, K.~Chakraborty, H.~Wang, and S.~Bansal, ``On safety and liveness filtering using hamilton–jacobi reachability analysis,'' \emph{IEEE Transactions on Robotics}, vol.~40, pp. 4235--4251, 2024.

\bibitem{lin2024onefilter}
\BIBentryALTinterwordspacing
A.~Lin, S.~Peng, and S.~Bansal, ``One filter to deploy them all: Robust safety for quadrupedal navigation in unknown environments,'' 2024. [Online]. Available: \url{https://arxiv.org/abs/2412.09989}
\BIBentrySTDinterwordspacing

\end{thebibliography}

\end{document}